\documentclass{article} 
\usepackage{natbib}

\usepackage{hyperref}
\usepackage{float}
\usepackage{verbatim} 
\usepackage{apalike}

\restylefloat{figure}
\restylefloat{table}

\usepackage{amsthm} 

\def\b{\boldsymbol} 

\usepackage{multicol}
\usepackage{multirow}
\usepackage{makecell}

\usepackage{siunitx}
\usepackage{booktabs}

\theoremstyle{definition}
\newtheorem{definition}{Definition}

\usepackage{mathtools}
\usepackage{relsize}
\usepackage{amssymb}
\usepackage{eurosym} 

\usepackage{color}
\usepackage{xcolor}
\definecolor{my_Blue}{HTML}{B7DEE8}
\definecolor{my_Green}{HTML}{D8E4C0}
\definecolor{my_Red}{HTML}{E6B9B5}
\definecolor{my_Purple}{HTML}{DDD7E6}
\definecolor{my_dark_Blue}{HTML}{4BACC6}
\definecolor{my_dark_Green}{HTML}{9DBB61}
\definecolor{my_dark_Red}{HTML}{C05046}
\definecolor{my_dark_Purple}{HTML}{AB9AC0}
\usepackage{tcolorbox}
\tcbuselibrary{theorems}

\newtcbtheorem[number within=section]{myalgo}{Algorithm}%
{colback=my_Red!20,colframe=my_dark_Red!100,fonttitle=\bfseries}{algo}

\usepackage{algorithm}
\usepackage{algpseudocode}

\usepackage{ulem} 

\providecommand{\keywords}[1]
{
  \small	
  \textbf{\textit{Keywords---}} #1
}

\usepackage{authblk}
\usepackage[a4paper, scale=0.72]{geometry}
\usepackage{hyperref}
\begin{document}

\title{Finding Money Launderers Using Heterogeneous Graph Neural Networks}

\date{}

\author[1]{Fredrik Johannessen\thanks{fredjo89@gmail.com}}
\author[2]{Martin Jullum\thanks{Corresponding author: jullum@nr.no}}
\affil[1]{DNB, P.O. Box 1600, Sentrum, NO-0021 Oslo, Norway}
\affil[2]{Norwegian Computing Center, P.O. Box 114, Blindern, NO-0314 Oslo, Norway}

\maketitle

\begin{abstract}
Current anti-money laundering (AML) systems, predominantly rule-based, exhibit notable shortcomings in efficiently and precisely detecting instances of money laundering. 
As a result, there has been a recent surge toward exploring alternative approaches, particularly those utilizing machine learning.
Since criminals often collaborate in their money laundering endeavors, accounting for diverse types of customer relations and links becomes crucial. 
In line with this, the present paper introduces a graph neural network (GNN) approach to identify money laundering activities within a large heterogeneous network constructed from real-world bank transactions and business role data belonging to DNB, Norway's largest bank.
Specifically, we extend the homogeneous GNN method known as the Message Passing Neural Network (MPNN) to operate effectively on a heterogeneous graph. As part of this procedure, we propose a novel method for aggregating messages across different edges of the graph. 
Our findings highlight the importance of using an appropriate GNN architecture when combining information in heterogeneous graphs.
The performance results of our model demonstrate great potential in enhancing the quality of electronic surveillance systems employed by banks to detect instances of money laundering. To the best of our knowledge, this is the first published work applying GNN on a large real-world heterogeneous network for anti-money laundering purposes.
\end{abstract}


\section{Introduction}
\label{introduction}

Money laundering is the activity of securing proceeds of a criminal act by concealing where the proceeds come from. The ultimate goal is to make it look like the proceeds originated from legitimate sources. 
Money laundering is a vast global problem, as it enables all types of crime where the goal is to make a profit. 
Because most money laundering goes undetected, it is difficult to quantify its effect on the global economy. 
However, a research report by \cite{UNODC} estimates that 1-2 trillion US dollars are being laundered each year, which corresponds to 2-5\% of global gross domestic product.
Both national and international anti-money laundering (AML) laws regulate electronic surveillance and reporting of suspicious transaction activities in financial institutions.
The purpose of the surveillance is to detect \textit{suspicious activities} with a high probability of being related to money laundering, such that they can be manually investigated.
The manual investigation is performed by experienced investigators, which inspect several aspects of the case, often involving multiple implicated customers, to then decide whether the behavior is suspicious enough to be reported to the authorities.
See Section \ref{sec:background_related_work} for more details about this process.
The present paper concerns the electronic surveillance process, which ought to identify a relatively small number of suspicious activities in a vast ocean of legitimate ones.

In the past few decades, the electronic surveillance systems in banks have typically consisted of several simple rules, created by domain experts, that use fixed thresholds and a moderate number of \texttt{if/else} statements that determine whether an alert is generated.
Such rules are still in large parts what makes up the surveillance systems \citep{CHEN2018}.  
However, these rules fall short of providing efficiency \citep{REUTERS}, for at least three key reasons: 
a) They rely on manual work to create and keep the rules up to date with the dynamically changing data \citep{CHEN2018}. This work increases with the complexity and number of rules. 
b) They are typically too simple to be able to detect money laundering with high precision, resulting in many low-quality alerts, i.e. \textit{false positives}\footnote{Hiring more workers to inspect each alert manually is often the resolution to compensate for this shortcoming.}.
c) Their simplicity makes them easy to circumvent for sophisticated money launderers, resulting in the possibility that severe crimes go undetected. 
The consequence of these three shortcomings is that banks spend huge resources on a very inefficient approach, and only a tiny fraction of illegal proceeds is being recovered \citep{POL2020}. 

Driven by multiple money laundering scandals in recent years\footnote{
As an example, it was revealed in 2018 that the Estonian branch of Danske Bank, Denmark's largest bank, carried out suspicious transactions for over \euro200 billion during 2007-2015 \citep{BJERREGAARD2019}. 
Subsequent lawsuit claims have amounted to over \euro2 billion.
}, 
the shortcomings of the rule-based system are high on the agenda for regulators and financial institutions alike, and considerable resources are devoted to developing more effective electronic surveillance systems. 
One avenue that is explored is to use machine learning (ML) to automatically learn when to generate alerts (see e.g.~\cite{JULLUM2020}, \cite{ROCHA2021}). 
Compared to human capabilities, ML is superior at detecting complicated patterns in vast volumes of data. 
As a consequence, ML-based systems have the potential to provide detection systems with increased accuracy and the ability to identify more sophisticated ways of laundering money.

Money laundering is a social phenomenon, where groups of organized criminals often collaborate to launder their criminally obtained proceeds. 
In the networks relevant for AML, the nodes may consist of customers, while the edges between the nodes represent money transfers, shared address, or joint ownership, to name a few possibilities. Analyses of such networks can uncover circumstances that would be impossible to detect through a purely entity-based approach, simply because the necessary information would not be present. 
Therefore, ML methods that leverage these relational data have a substantial advantage over those that do not. 
The simplest way to incorporate network characteristics into ML methods is
to create node features that capture information about the node's role in the network, e.g.~network centrality metrics  or characteristics of its neighborhood.
The features can then be used in a downstream machine learning task. 
This approach is, however, suboptimal as the generated features might not be the ones most informative for the subsequent classification, and the full richness of the relational data will in any case not be passed over to the entity-based machine learning task.

\textit{Graph Neural Network} (GNN) is a class of methods that overcome this drawback by applying the machine learning task directly on the network data through a neural network. 
GNNs are able to solve various graph-related tasks such as node classification \citep{KIPF2016}, link prediction \citep{ZHANG2018}, graph clustering \citep{WANG2017}, graph similarity \citep{BAI2019} as well as unsupervised embedding \citep{KIPF2016variational}. 
The primary idea behind GNNs is to extend the modern and successful techniques of \textit{artificial neural networks} (ANNs) from regular tabular data to that of networks. 
In addition to their ability to incorporate both entity features and network features into a single, simultaneously trained model, most GNNs scale linearly with the number of edges in the network, making them applicable to large networks.
A huge benefit of GNNs in practical use cases is that they are inductive rather than transductive:  While transductive models (e.g.~Node2Vec \citep{GROVER2016}) can only be used on the specific data that was present during training, inductive models can be applied to entirely new data without having to be retrained. 
This is crucial for the AML application where new transactions (edges) appear continuously, and customers (nodes) enter and leave on a daily basis. 

Most GNNs are developed for \textit{homogeneous} networks with a single type of entity (node) and a single type of relation (edge).
In the present AML use case, the relevant network is \textit{heterogeneous} in both nodes and edges: The nodes represent both private customers, companies, and external accounts, while the edges represent both financial transactions and professional business ties between the nodes. 
There exist some GNNs in the literature that are able to handle heterogeneous networks, such as:
RGCN \citep{SCHLICHTKRULL2018},
HAN \citep{WANG2019},
MAGNN \citep{FU2020MAGNN},
HGT \citep{HU2020HGT}, and 
HetGNN \citep{ZHANG2019HetGNN}.
A common issue with all these methods is that they are not designed to incorporate edge features. 
In our AML use case, this corresponds to properties of the financial transactions (or the business ties) and is crucial information for an effective learning task.
Thus, based on our current knowledge and research, there exists no directly applicable GNN method for our AML use case.

The present paper proposes a heterogeneous GNN method that utilizes the edge features in the graph, which we denote \textit{Heterogeneous Message Passing Neural Network} (HMPNN). 
The HMPNN method is an expansion of the MPNN method \citep{GILMER2017} to a heterogeneous setting.
The extension essentially connects, and simultaneously trains multiple MPNN architectures, each working for different combinations of node and edge types.
We investigate two distinct approaches to aggregate the embeddings of node-edge combinations in the final step: 
The first approach applies a simple summation of the embedding vectors derived from the various combinations. 
The second approach is novel and concatenates the embeddings before applying an additional single-layer neural network to enhance the aggregation process.

The HMPNN is developed for and applied to detect money laundering activities in a large heterogeneous network created from real-world bank transactions and business role data which belongs to Norway's largest bank, DNB. 
The nodes represent bank customers and transaction counterparties, while the edges represent bank transactions and business ties.
The network has a total of more than 5 million nodes and almost 10 million edges. 
Among the bank customers, some are known to conduct suspicious behavior related to money laundering. While the rest are assumed not fraudulent, there could be some undetected suspicious behavior also there. Thus, this is actually a semi-labeled dataset. 
Finally, no oversampling or undersampling was performed, which makes the class imbalance realistic to what one encounters in practical situations. 

There are primarily two categories of bank customers: individual (retail) customers and organization (corporate) customers. 
Due to the fundamental differences between these two groups and their fraudulent behaviors, it is common practice to study and model their fraudulent behavior separately. 
In our setting, we, therefore, treat them as two distinct node types, each with its unique set of features.
In this paper, we limit the scope to modeling, predicting, and detection of fraudulent \textit{individual} customers. 
I.e.~all fraudulent customers in our dataset are individual customers. 
We chose to restrict our analysis to individuals because there is a significantly larger number of known fraudulent individuals compared to organizations, providing the model with more fraudulent behavior to learn from.
Furthermore, individuals are a more homogeneous group with simpler customer relationships compared to organizations, making them a more suitable group to apply our methodology to. 
Nevertheless, note that organizations may very well be involved and utilized by fraudulent individuals, and possess key connections in the graph that the model learns from.

In addition to network data, the nodes and edges have sets of features that depend on what type of node and edge they are. 
In this paper, HMPNN is compared to other state-of-the-art GNN methods and achieves superior results. 
To the best of our knowledge, there exists no prior published work on applying graph neural networks on a large real-world heterogeneous network for the purpose of AML.

The rest of this article is organized as follows: 
Section \ref{sec:background_related_work} provides some background and an overview of related work within the AML domain, as well as the GNN domain.
In Section \ref{sec:model} we formulate our HMPNN model after introducing the necessary mathematical framework and notation. 
Section \ref{sec:usecase} presents the data for our AML use case, the setup of our experiment, and presents and discusses the results we obtain.
Finally, Section \ref{sec:conclusion} summarizes our contribution, and provides some concluding remarks and directions for further work.
Additional model details are provided in Appendix\ref{app:network_features} and \ref{app:parameters}.

\section{Background and related work}
\label{sec:background_related_work}

In this section, we provide some background on the AML process and give an overview of earlier, related work in the domain of AML. 
We also briefly describe the state-of-the-art of homogeneous and heterogeneous GNNs in the general case.

\subsection{AML process}

\begin{figure}[t]
\centering
\includegraphics[width=0.9\linewidth]{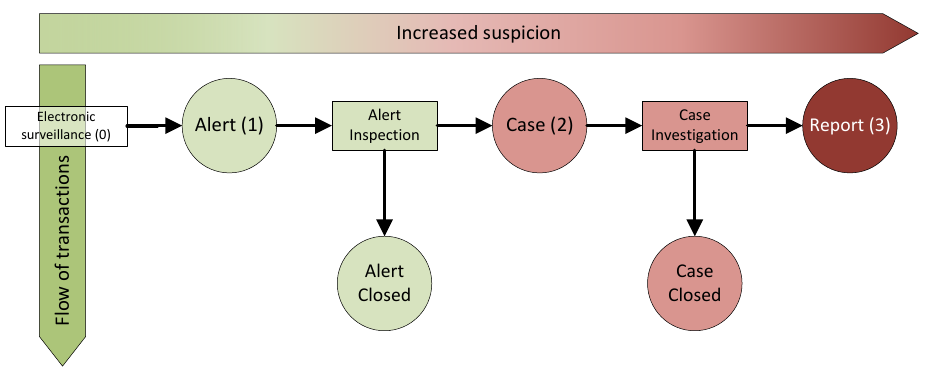}
\caption{Workflow following detections from the electronic surveillance system.}
\label{fig:alert_workflow}
\end{figure}

As mentioned in the Introduction, financial institutions are required by law to have effective AML systems in place. 
Figure \ref{fig:alert_workflow} illustrates a typical AML process in a bank.
The electronic surveillance system generates what is called \textit{alerts} (1) on some of the transactions, or transaction patterns, that go through the bank.  
An alert is first subject to a brief initial inspection, where those that can easily be identified as legitimate are picked out and marked as closed.  
Otherwise, the alert is upgraded to a \textit{case} (2).  
At this stage, multiple alerts might be merged into a single case, which regularly involves multiple implicated customers. 
The case is then inspected thoroughly by experienced investigators.  
If money laundering is ruled out, the case will be marked as closed. 
Otherwise, the case will be upgraded to a \textit{report} (3), and the revealed circumstances are reported in detail to the national \textit{Financial Intelligence Unit} (FIU). 
From here on, the FIU oversees further action and will determine whether to start a criminal investigation. 

The manual investigation part of the process is difficult to set aside for two reasons: a) The process is hard to automate, as the investigators sit on crucial, yearlong experience and often use non-quantifiable information sources to build their cases. b) AML laws do typically not allow automated reporting of suspicious behavior. 
The electronic surveillance generating the alerts is suitable for automation, and the benefits of better and more targeted alerts with fewer false positives, are huge for financial institutions. 
That is also why this paper is concerned with the electronic surveillance aspect of the AML process.

\subsection{AML literature}

In AML literature, there are very few papers that have datasets with real money laundering cases, and the majority of articles are validated on small datasets consisting of less than 10,000 observations \citep{CHEN2018}. 

Both supervised (\cite{LIU2008}, \cite{DENG2009}, \cite{ZHANG2019}, \cite{JULLUM2020}) and unsupervised  (\cite{LORENZ2020}, \cite{ROCHA2021}) machine learning have been applied to AML use cases. However, the literature is quite limited, particularly for papers utilizing relational data. 

Some AML papers apply the aforementioned strategy of generating network features used in a down-stream machine learning task:
\cite{SAVAGE2016} create a graph from cash- and international transactions reported to the Australian FIU (AUSTRAC), from which they extract communities defined by filtered $k$-step neighborhoods around each node, to then create summarizing features used to classify community suspiciousness using supervised learning.
\cite{COLLADON2017} construct multiple graphs (each of which aims to reveal unique aspects) from customer and transaction data belonging to an Italian factoring company. On these graphs, they report significant correlation between traditional centrality metrics (e.g. betweenness centrality) and known high-risk entities.
\cite{ELLIOTT2019} uses a combination of network comparison and spectral analysis to create features that are applied in a downstream learning algorithm to classify anomalies. 

There are only a few papers applying GNNs to money laundering detection:
In a brief paper, \cite{weber2018} discuss the use of GNN on the AML use case.
The paper provides some initial results of the scalability of GCN \citep{KIPF2016} and FastGCN \citep{chen2018fastgcn} for a synthetic data set. 
However, no results on the performance of the methods were provided. 
\cite{weber2019} compare GCN to other non-relational ML methods on a real-world graph generated from bitcoin transactions, with 203k nodes and 234k edges. 
The authors highlight the usefulness of the graph data and find that GCN outperforms logistic regression, but it is still outperformed by random forest. 
The dataset, called the Elliptic data, is also released with the paper and has later been utilized by several others: 
\cite{alarab2020competence} apply GCN extended with linear layers and improve significantly on the performance of the GCN model in \cite{weber2019}. 
\cite{lo2023inspection} apply a self-supervised GNN approach to create node embedding subsequently used as input in a random forest model, and reported good performance.
Others \citep{alarab2022graph, pareja2020evolvegcn} exploited the temporal aspect of the graph to increase the performance on the same data set. 

It is unknown to what extent results from synthetic or bitcoin transaction networks are transferable to the real-life application of transaction monitoring of bank transactions. Our graph is also about 25 times larger than the Elliptic dataset.
Moreover, as we will review immediately below, much has happened in the field of GNN since the introduction of GCN in 2018. 
Finally, to the best of our knowledge, there are no papers applying GNN (or other methodology) to an AML use case with a \textit{heterogeneous} graph. 

\subsection{General case GNN literature}

The history of GNNs can be traced back about twenty years and GNNs have during the past few years surged in popularity. 
This was kicked off by \citet{KIPF2016} who introduced the popular method \textit{Graph Convolutional Network} (GCN). 
For an excellent survey of GNNs including their history, we refer to \cite{WU2020}.
The core dynamics in GNNs is an iterative approach where each node receives information from its neighbors, and combines it with its own representation to create a new representation, which will be forwarded to its neighbors in the next iteration. 
We call this \textit{message passing}.
After a few iterations, these node representations are used to make inference about the node. 

Concentrating on today's most popular group of GNNs, \textit{Spatial-based convolutional GNNs}, we briefly review some relevant homogeneous and heterogeneous GNN methods below.

\subsubsection{Homogeneous GNN literature}
\label{sec:homGNN}

The GCN method \citep{KIPF2016} is motivated by spectral convolution and was originally formulated for the transductive setting operating directly on the adjacency matrix of the graph. However, the method can be reformulated in the inductive and spatial-based GNN setting using a message-passing formulation:  The message received by a node from its neighbors is a weighted linear transformation of the neighboring node representations, followed by aggregating the result by taking their sum.

GraphSage \citep{HAMILTON2017} expands on GCN in two ways: It uses a) a neighborhood sampling strategy to increase efficiency, and b) a neutral network, a pooling layer, and an LSTM \citep{hochreiter1997long} to aggregate the incoming messages instead of a simple sum. A drawback of GraphSage is, however, that it does not incorporate edge weights.

The Graph attention network (GAT) of \citet{VELIVCKOVIC2017} introduced the attention mechanism into the GNN framework. This mechanism learns the relative importance of a node's neighbors, to assign more weight to the most important ones.

\citet{GILMER2017} presented the Message Passing Neural Network (MPNN) framework, unifying a large group of different GNN models.
Apart from the unifying framework, the most essential contribution of this model is in our view that it applies a learned message-passing function that utilizes edge features.


\subsubsection{Heterogeneous GNN literature}
\label{sec:hetGNN}
Heterogeneous graph neural networks are commonly defined as extensions of existing homogeneous graph neural networks. For instance, the Relational Graph Convolution Network (RGCN) \citep{SCHLICHTKRULL2018} extends the GCN framework to support graphs with multiple types of edges. RGCN achieves this by breaking down the heterogeneous graph into multiple homogeneous ones, one for each edge type. In each layer, GCN is applied to each homogeneous graph, and the resulting node embeddings are element-wise summed to form the final output. A drawback of RGCN is that it does not take node heterogeneity into account.
Heterogeneous Graph Attention Networks (HAN) \citep{WANG2019} generalize the Graph Attention Network (GAT) approach to heterogeneous graphs by considering messages between nodes connected by so-called meta-paths. Meta-paths (see the formal definition in Definition \ref{def:meta_path_step}) are composite relationships between nodes that help to capture the rich structural information of heterogeneous graphs. 
HAN defines two sets of attention mechanisms. The first is between two different nodes, which is analogous to GAT. The second set of attention mechanisms is performed at the level of meta-paths, which computes the importance score of different composite relationships.
Metapath Aggregated Graph Neural Network (MAGNN) \citep{FU2020MAGNN} extends the approach of HAN by also considering the intermediary nodes along each meta-path \citep{DONG2017}. While HAN computes node-wise attention coefficients by only considering the features of the nodes at each end of the meta-path, MAGNN transforms all node features along the path into a single vector. 
%
%

The interest in GNNs is rapidly growing, and advancements in this field are consistently being made.
There are several other methods available that haven't been discussed here. 
For a more comprehensive understanding, we once again refer to the survey conducted by \cite{WU2020}, which provides an overview of GNNs in general. Additionally, for insights specifically on Heterogeneous Network Representation Learning, including Heterogeneous GNNs, we refer to \cite{YANG2020} for a good overview.
\section{Model}
\label{sec:model}
In this section we define and describe our proposed heterogeneous GNN model, which is based on the generic framework from Message Passing Neural Network (MPNN) introduced in \cite{GILMER2017}.
We start by introducing the original (homogeneous) MPNN model and algorithm before we move on to give precise definitions for our heterogeneous graph setup and present our novel extension of the MPNN model for heterogeneous networks.

\subsection{Message Passing Neural Network (MPNN)}
\label{sec:MPNN}

\cite{GILMER2017} introduces the generic MPNN framework which is able to express a large group of different GNN models, including GCN, GraphSage, and GAT. 
This is done by formulating the message passing with two learned functions, $M_k(\cdot)$, called the \textit{message functions}, and $U_k(\cdot)$, called the \textit{node updated functions}, with forms to be specified later.
The framework runs $K$ message passing iterations $k=1,\ldots,K$ between nodes along the edges that connect them. 
The node representation vectors are initialized as their feature vectors, $\b{h}_v^{0} = \b{x}_v$, and the previous representation $\b{h}_v^{(k-1)}$ is the message that is being sent in each iteration $k$.
After $K$ iterations, the final representation $\b{h}_v^{(K)}$ is passed on to an output layer to perform e.g.~node-level prediction tasks.
The message-passing function is defined as 
\begin{equation}\label{eq_mpnn}
\begin{aligned}
    \b{m}_v^{(k)} &= \sum_{u\in N(v)} M_k(\b{h}_v^{(k-1)}, \b{h}_u^{(k-1)}, \b{r}_{uv}), \\
    \b{h}_v^{(k)} &= U_k( \b{h}_v^{(k-1)}, \b{m}_v^{(k)}),
\end{aligned}
\end{equation}
where $N(v)$ denotes the neighborhood of node $v$, and $\b{r}_{uv}$ represents the edge features for the edge between node $u$ and $v$.
By defining specific forms of $U_k(\cdot)$ and $M_k(\cdot)$, a distinct GNN method is formulated. 
From the viewpoint of this paper, an essential attribute of the MPNN framework is that the learned message-passing function utilizes edge features.
Not many other proposed GNNs incorporate edge features into their model. 
\cite{GILMER2017} emphasize the importance of edge features in the dataset they experiment on. 
For our dataset, the edge features contain essential information related to transactions and are required to be utilized in an expressive model.

\subsection{Formal definitions}

Before we can formulate our \textit{heterogeneous} MPNN framework, we need to establish a precise definition of a heterogeneous graph, as well as a couple of additional concepts. 

We largely adopt the commonly used graph notation from \cite{WU2020}, 
and use a heterogeneous graph definition which is a slight modification to those in \cite{YANG2020} and \cite{WANG2019} in order to allow for multiple edges of different types between the same two nodes.

\begin{definition}(Heterogeneous graph)
\label{def:heteroGraph}
A heterogeneous graph is represented as $G = (V, E, \boldsymbol{X}, \boldsymbol{R}, Q^V, Q^E, \phi)$ where each node $v\in V$ and each edge $e\in E$ has a type, and $Q^V$ and $Q^E$ denote finite sets of predefined node types and edge types, respectively.
Each node $v\in V$ has a node type $\phi(v) = \nu \in Q^V$, where $\phi(\cdot)$ is a node type mapping function. 
Further, for $\phi(v) = \nu$, $v$ has features $\b{x}_v^{\nu} \in \b{X}^{\nu}$, 
where 
$\b{X}^{\nu} = \{ \b{x}_v^{\nu}\mid v \in V,\, \phi(v) = \nu \}$ and
$\b{X} = \{\b{X}^{\nu} \mid \nu \in Q^V\}$.
The dimension and specifications of the node feature $\b{x}_v^{\nu}$ may be different for different node types $\nu$.
Further, let us denote by $e_{uv}^{\varepsilon}$ an edge of type $\varepsilon \in Q^E$ pointing from node $u$ to $v$. 
Each edge $e_{uv}^{\varepsilon}$ has features 
$\b{r}_{uv}^{\varepsilon} \in \b{R}^{\varepsilon}$, where 
$\b{R}^{\varepsilon} = \{\b{r}_{uv}^{\varepsilon} \mid u,v\in V \}$ and
$\b{R} = \{ \b{R}^{\varepsilon} \mid  \varepsilon \in Q^E\}$. 
Just like for nodes, the edge features may have different dimensions for different edge types.
\end{definition}

To formulate heterogeneous message passing, we will use the concept of \textit{meta-paths}. Meta-paths are commonly used to extend methods from a homogeneous to a heterogeneous graph. For example, \cite{DONG2017} use meta-paths when introducing \textit{metapath2vec}, which extends \textit{DeepWalk} \citep{PEROZZI2014} and the closely related \textit{node2vec} \citep{GROVER2016} to a method applicable on heterogeneous graphs.
\cite{WANG2019} use meta-paths to generalize the approach of graph attention networks \citep{VELIVCKOVIC2017} to that of heterogeneous graphs when formulating \textit{heterogeneous graph attention network} (HAN).
In addition to meta-path, the below definition introduces our own term, \textit{meta-steps}, which we use when formulating our model. 

\begin{definition}(Meta-path, meta-step)
\label{def:meta_path_step}
A \textit{Meta-path} belonging to a heterogeneous graph $G$ is a sequence of specific edge types between specific node types, 
\[
    (\nu_0, \varepsilon_1, \nu_1, \varepsilon_2, \dots,\nu_{k-1}, \varepsilon_k, \nu_{k}), \quad \nu_i \in Q^V , \varepsilon_i \in Q^E.
\]
Here, $k$ is the length of the meta-path.
Let $S$ be the set of meta-paths of length 1, 
\[
    S = \{ s=(\mu,\varepsilon,\nu) \mid  \mu,\nu \in Q^V, \varepsilon \in Q^E \},
\]
and refer to the elements $s\in S$ as \textit{meta-steps}.
\end{definition}

Finally, we introduce a definition of node neighborhood over a specific meta-step:
\begin{definition}(Meta-step specific node neighborhood)
Let $N_{\mu}^{\varepsilon}(v)$ be the set of nodes of type $\mu$ which is connected to node $v$ by an edge of type $\varepsilon$ pointing to $v$:
\[
N_{\mu}^{\varepsilon}(v) = 
\{ u\in V \mid \phi(u) = \mu, e_{uv}^{\varepsilon}\in E   \}.
\]
We call $N_{\mu}^{\varepsilon}(v)$ the (incoming) node neighborhood to node $v$ with respect to the meta-step $s = (\mu, \varepsilon, \phi(v)) \in S$.
\end{definition}

\subsection{Heterogeneous MPNN}
\label{sec:HMPNN}
We are now ready to formulate our heterogeneous version of the MPNN method (HMPNN). 
The complete algorithm is provided in Algorithm \ref{alg:HMPNN}.

Our approach for extending a homogeneous GNN to a heterogeneous one is, essentially, the same as used by \cite{SCHLICHTKRULL2018}, where they generalize GCN to the method \textit{Relational Graph Convolutional Network} (RGCN) applicable on graphs with multiple edge types. 

The algorithm performs (at each iteration) multiple MPNN message passing operations, one for each meta-step $s \in S$ in the graph. Each of these has its separate learned functions $M_k^s(\cdot) = M_k^{(\mu,\varepsilon,\nu)}(\cdot)$ and $U_k^s(\cdot) = U_k^{(\mu,\varepsilon,\nu)}(\cdot)$, which allows the method to learn the context of each meta-step, and also allow message passing between nodes and across edges with varying numbers of features. 
The intermediate output of this process is multiple representation vectors for each node. To reduce these to a single vector, they are aggregated by a learned \text{aggregation function} $A^{\nu}(\cdot)$ which is specific to each node type. In line with the generic formulation of MPNN we do not specify a specific form of the aggregation function in Algorithm \ref{alg:HMPNN}. During the experiments, we have assessed two alternative options, which we will discuss shortly.

Our HMPNN model is implemented in Python, using the library PyTorch Geometric (PyG) \citep{Fey2019FastGR} allowing for high-performance computing by utilizing massive parallelization through GPUs. 
Source code is available here: \url{https://github.com/fredjo89/heterogeneous-mpnn}

\begin{algorithm}[ht!]
\caption{Heterogeneous MPNN}\label{alg:HMPNN}
\footnotesize
\begin{algorithmic}
\State Initialize the representation of each node as its feature vector,
\[
\b{h}_v^{(0)} = \b{x}_v^{\nu}, \;\forall\; v \in V \text{ of type } \nu = \phi(v).
\]

\For{$k = 1,\dots, K$} \Comment{For each iteration}
\For{$v \in V$} \Comment{For each node}
\For{$\mu \in Q^V, \varepsilon \in Q^E$ such that $N_{\mu}^{\varepsilon}(v) \neq \emptyset$} \Comment{For each meta-step ending at $\phi(v)$}

\State Compute the new representation for node $v$ of type $\nu=\phi(v)$, for the specific
\State meta-step,
\[
\begin{aligned}
    \b{m}_v^{(\mu, \varepsilon,k)} &= \sum_{u\in N_{\mu}^{\varepsilon}(v)} M_k^{(\mu, \varepsilon, \nu)}(\b{h}_v^{(k-1)}, \b{h}_u^{(k-1)}, \b{r}_{uv}^{\varepsilon}), \\
    \b{h}_v^{(\mu, \varepsilon,k)} &= U_k^{(\mu, \varepsilon, \nu)}( \b{h}_v^{(k-1)}, \b{m}_v^{(\mu, \varepsilon,k)}).
\end{aligned}
\]
\EndFor
\State Aggregate the representations from the multiple meta-steps into a 
\State single new representation for that node,
\State
\[
    \b{h}_v^{(k)} = A_k^{(\nu)}\big( 
    \{
    \b{h}_v^{(\mu, \varepsilon,k)} \mid \mu \in Q^V, \varepsilon \in Q^E
    \}  
    \big)
\]
\EndFor

\EndFor
\end{algorithmic}
\end{algorithm}

\subsection{Choosing specific forms of the functions}

As our setup is very general, the algorithm in Algorithm \ref{alg:HMPNN} gives rise to a whole range of new heterogeneous GNN methods. 
By defining specific forms of $U_k^{(\mu, \varepsilon, \nu)}(\cdot)$, $M_k^{(\mu, \varepsilon, \nu)}(\cdot)$ and $A_k^{(\nu)}(\cdot)$, a distinct GNN method is formulated. 
Note that there is nothing that prevents us from choosing different forms of these functions for different meta-steps or iterations. 
However, for the AML use case in Section \ref{sec:usecase}  we have limited the scope to a single form for each of the three functions, respectively. These are described in the following.

As message function, we use the same as \cite{GILMER2017}:
\[
M_k^{(\mu, \varepsilon, \nu)}(\b{h}_v^{(k-1)}, \b{h}_u^{(k-1)}, \b{r}_{uv}^{\varepsilon}) = g_k^{(\mu, \varepsilon, \nu)}(\b{r}_{uv}^{\varepsilon}) \b{h}_u^{(k-1)}.
\]
Here, $g_k^{(\mu, \varepsilon, \nu)}(\cdot)$ is a single layer neural network which maps the edge feature vector $\b{r}_{uv}^{\varepsilon}$ to a $d^v\times d^u$ matrix, where $d^u$, and $d^v$ are the number of features for the sending and receiving node type, respectively. 

As update function we use 
\[
U_k^{(\mu, \varepsilon, \nu)}\big(\b{h}_v^{(k-1)}, \b{m}_v^{(\mu, \varepsilon, \nu,k)}\big) = \sigma\Big(\b{m}_v^{(\mu, \varepsilon,k)} + \b{B}_k^{(\mu, \varepsilon, \nu)} \b{h}_v^{(k-1)}\Big),
\]
where $\b{B}_k^{(\mu, \varepsilon, \nu)}$ is a matrix.
Note that for a homogeneous graph, our choices of $M(\cdot)$ and $U(\cdot)$ are similar to those in \cite{HAMILTON2017}, except that the matrix applied in the message function is conditioned on the edge features rather than being the same across all edges. 

\begin{figure}[t]
\centering
\includegraphics[width=0.7\linewidth]{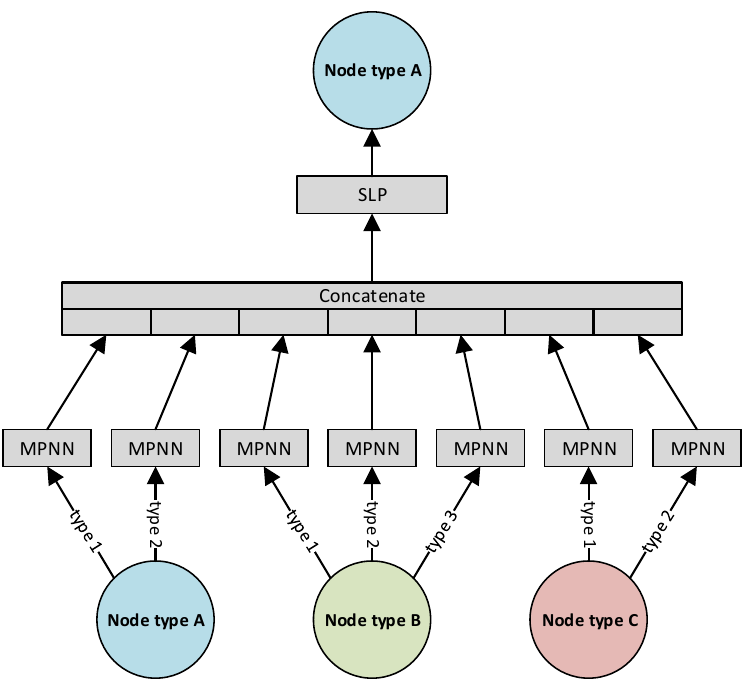}
\caption{Illustration of the architectural extension of the homogeneous MPNN model to our HMPNN model (HMPNN-ct) as messages are passed to one of the node types (A), from three node types (A, B, and C) and with three edge types (1, 2 and 3). SLP is short for Single Layer Perceptron (neural network). Analogous architectures are used for messages passed to the other node types.
}
\label{fig:MPNN-architecture}
\end{figure}

For the aggregation function, we consider two alternatives. The first is to take the sum of the vectors from each meta-step before performing a nonlinear transformation, in the same fashion as \citep{SCHLICHTKRULL2018}:
\begin{equation}\label{eq_mpnn_agg_sum}
A_k^{(\nu)}\big( 
    \{
    \b{h}_v^{(\mu, \varepsilon,k)} \mid \mu \in Q^V, \varepsilon \in Q^E
    \}  
    \big)
=
\sigma\bigg(
\underset{\mu \in Q^V, \varepsilon \in Q^E}{\sum} 
\b{h}_v^{(\mu, \varepsilon,k)}
\bigg).
\end{equation}
Here, $\sigma(\cdot)$ is the sigmoid function.
In the second aggregation method, the vectors $\b{h}_v^{(\mu,\varepsilon,k)}$ 
are concatenated into a single vector. A single-layer perceptron (neural network) is then applied to it, and outputs the new representation:
\begin{equation}\label{eq_mpnn_agg_ct}
A_k^{(\nu)}\big( 
    \{
    \b{h}_v^{(\mu, \varepsilon,k)} \mid \mu \in Q^V, \varepsilon \in Q^E
    \}  
    \big)
=
\sigma\Big(
W_k^{(\nu)}
\underset{\mu \in Q^V, \varepsilon \in Q^E}{||} 
\sigma \big(\b{h}_v^{(\mu, \varepsilon,k)}\big)
\Big).
\end{equation}
We denote the two resulting models \textit{HMPNN-sum} and \textit{HMPNN-ct}, respectively.
Figure \ref{fig:MPNN-architecture} illustrates the architectural extension of the homogeneous MPNN model to our HMPNN model (HMPNN-ct) as messages are passed to one of the node types, where \eqref{eq_mpnn_agg_ct} is used as aggregation function. 

\section{AML use case}
\label{sec:usecase}

In this section, we first describe the heterogeneous graph data. Then we lay out the setup of the experiments we have performed on this dataset before we provide the results.  

\subsection{Data}

The graph is established based on customer and transaction data from Norway's largest bank, DNB, in the period from February 1, 2022, to January 31, 2023.
The nodes in the graph represent entities that are senders and/or recipients of financial transactions.
If two entities participate in a transaction with each other, this is represented by an edge (of type transaction) between the two, where the direction of the edge points from the sender to the recipient.
There are in total 5 million nodes and 9 million such edges in the graph. 

There are three types of nodes in the graph. 
The first one is called \textit{individual} and represents a human individual’s customer relationship in the bank. 
It includes all of the individual’s accounts in the bank\footnote{ 
Transactions made to/from any of the accounts in the bank belonging to the customer will result in an edge to/from the node representing the individual. 
}.
The second type of node is called \textit{organization}, and represents an organization’s or company’s customer relationship in the bank in the same manner as a node representing an individual. 
The third type of node is called \textit{external} and represents a sender or recipient of a transaction that is outside of the bank. 

\begin{figure}[!t]
\centering
\includegraphics[width=0.6\linewidth]{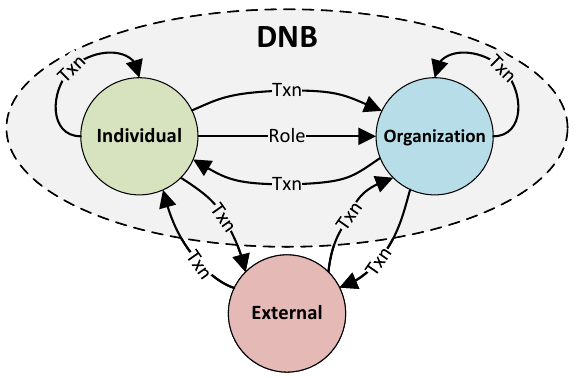}
\caption{The schema of the graph that has been the subject of our experiments, with the nodes representing customer relationships in DNB outlined by the grey ellipse. Here, the transaction edges are abbreviated as \textit{Txn}. }
\label{fig:graph_schema}
\end{figure}

The majority of the edges in the graph represent presence of a financial transaction between different individuals/organizations/external entities in the edge direction. 
In addition to this edge type, the graph includes role as a second edge type.
That edge points from an individual to an organization if the individual occupies a position on the board, is the CEO, or holds ownership in the organization. 
The resulting graph is directed and heterogeneous with respect to both nodes and edges. 
Figure \ref{fig:graph_schema} shows the schema of the graph, including the nine possible meta-steps. 
As shown in the schema, there are no edges between different external nodes,
since the bank does not have access to transactions not involving their customers. 

The nodes that represent individuals are assigned a binary class (0 for regular individuals, and 1 for individuals known to conduct suspicious behavior). 
As mentioned in the introduction, the data only contains labels for individual nodes.
Suspicious individuals are defined as those that have been subject of an AML case (stage 2 
in Figure \ref{fig:alert_workflow}) during a certain time window. 
Note that customers implicated in cases that were not reported to the FIU are still defined as suspicious. This decision was made because our objective is to model suspicious activity, which these customers certainly have conducted, even though the suspiciousness was diminished by a close manual inspection.
Less than 0.5\% of the individuals belong to class 1 (suspicious).

\begin{figure}[!t]
\centering
\includegraphics[width=0.49\linewidth]{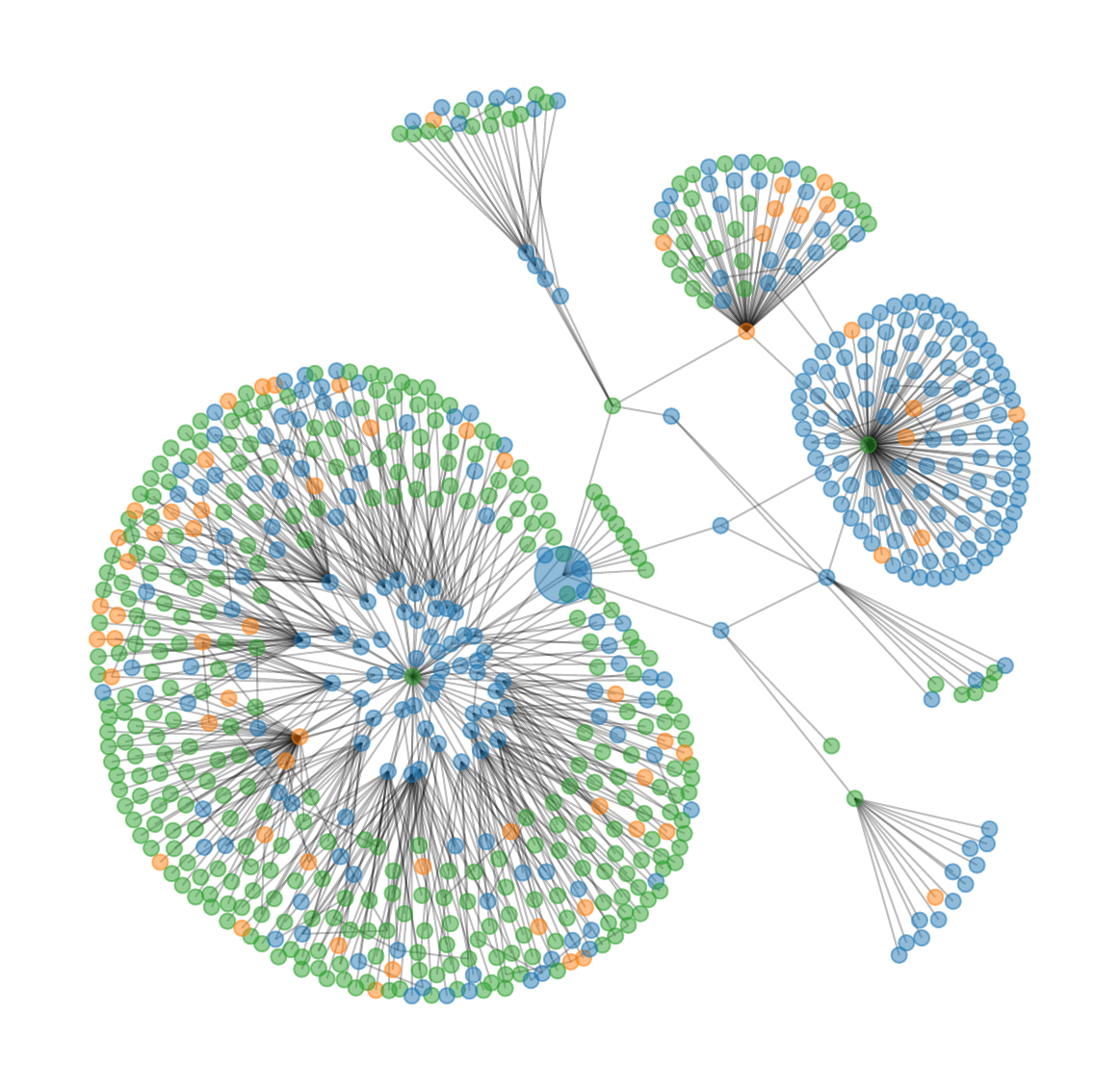}
\includegraphics[width=0.49\linewidth]{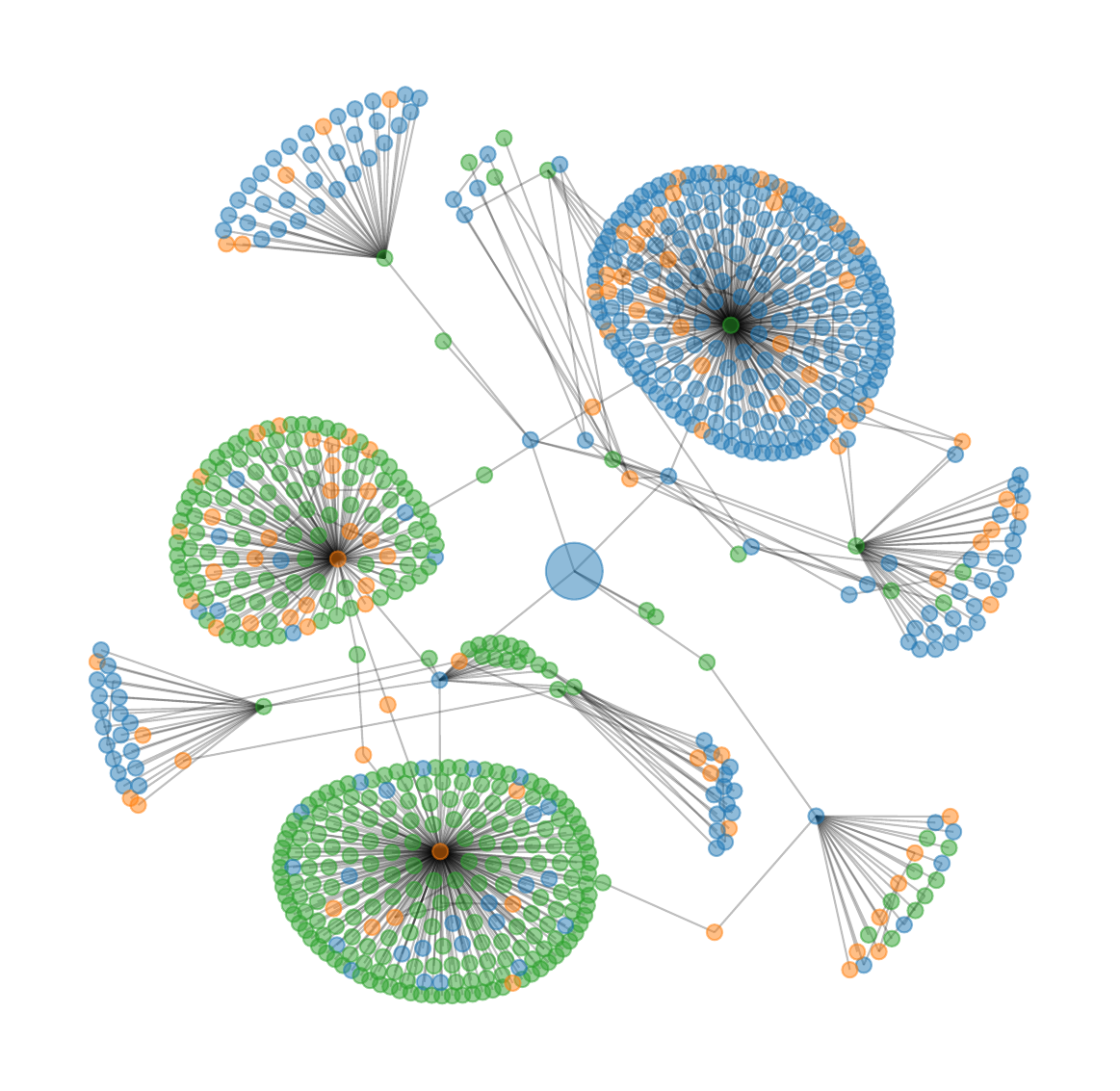}\\
\includegraphics[width=0.49\linewidth]{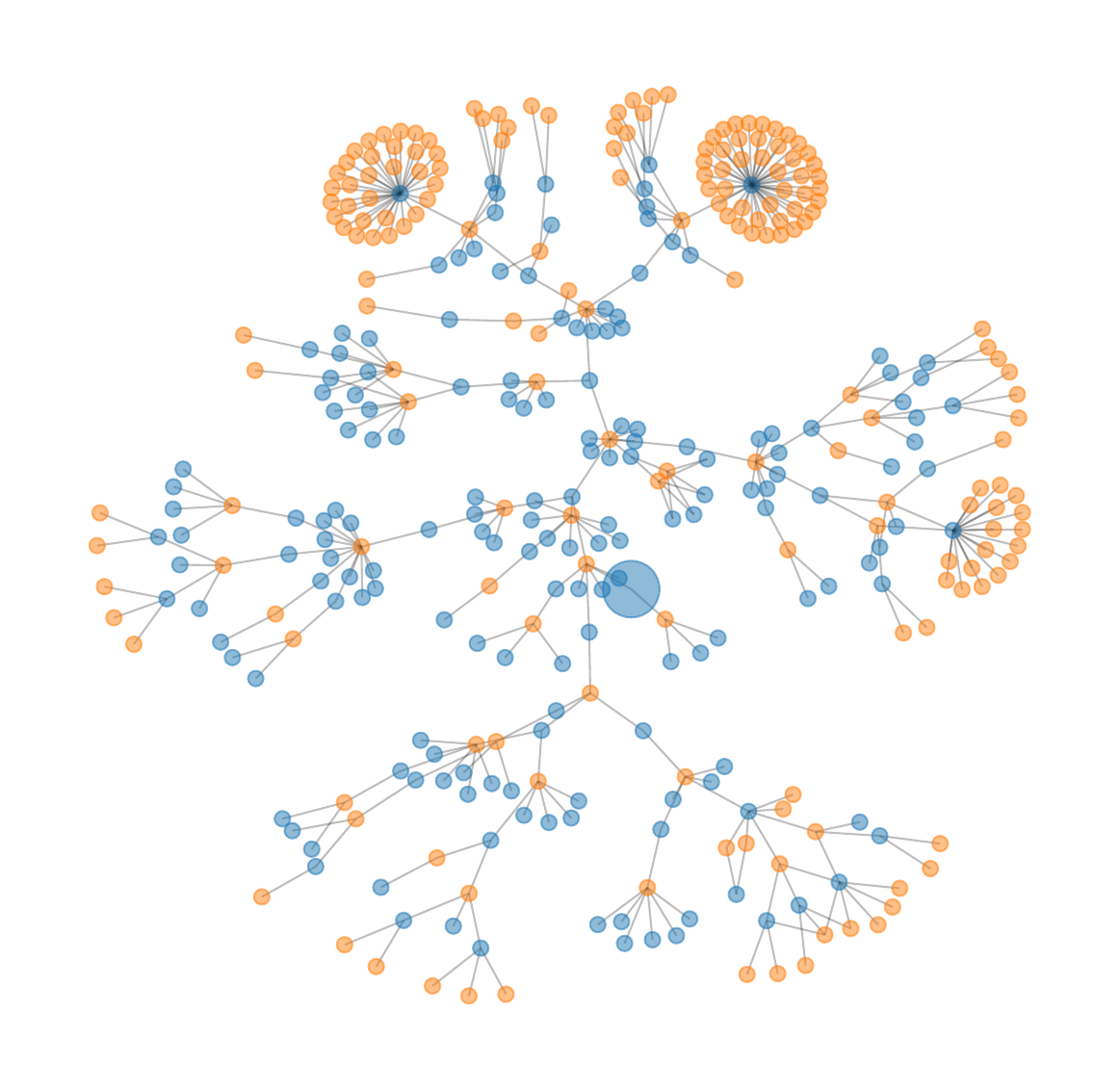}
\includegraphics[width=0.49\linewidth]{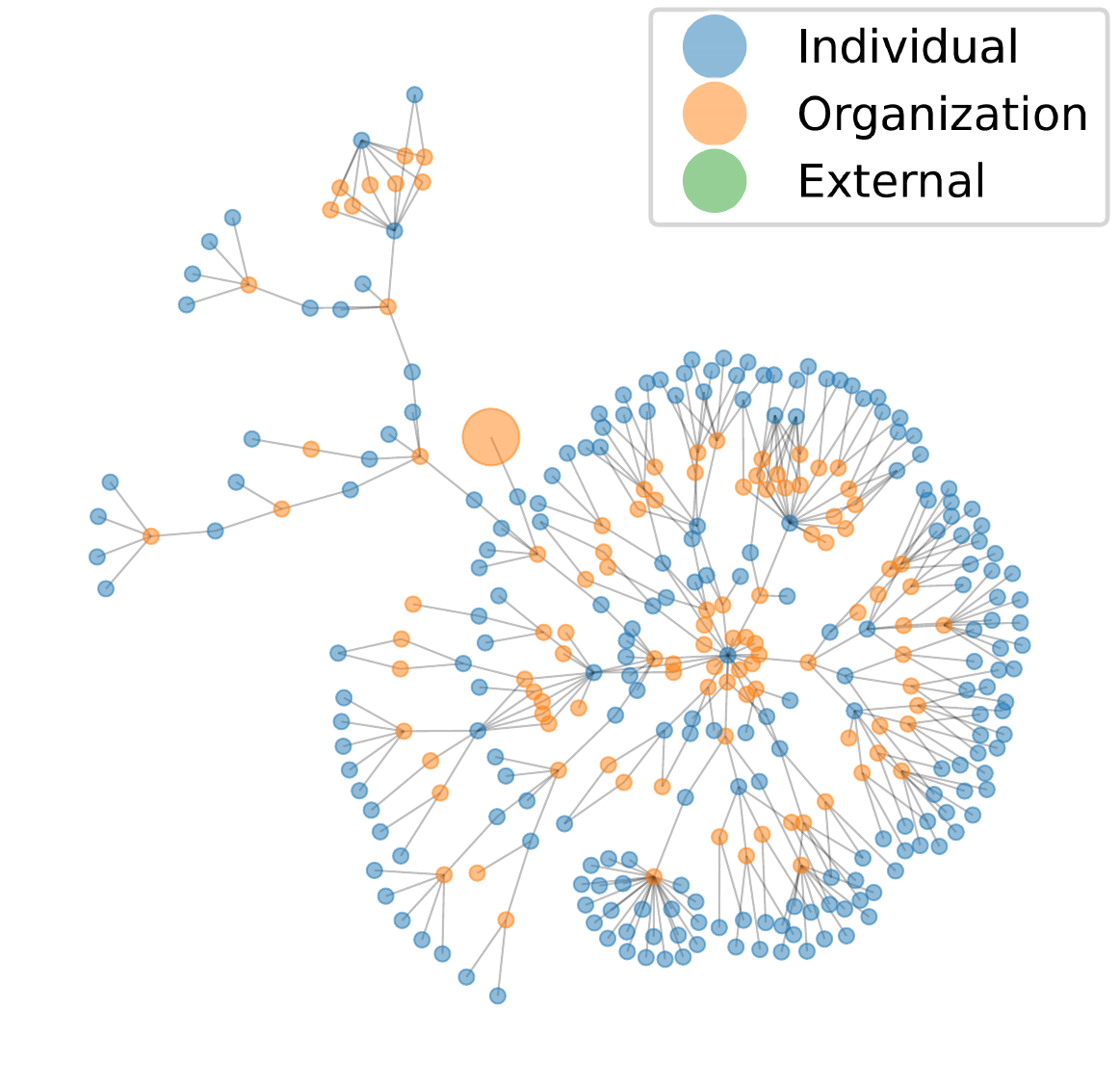}
\caption{Homogeneous egonets for four random nodes in the graph. 
The upper two panels show 3-hop egonets of the (undirected) transaction edges in the graph. 
The lower two panels show 9-hop egonets for the (undirected) role edges in the graph.
The starting node is enlarged.}
\label{fig:egonet}
\end{figure}

\begin{figure}[!t]
\centering
\includegraphics[width=0.99\linewidth]{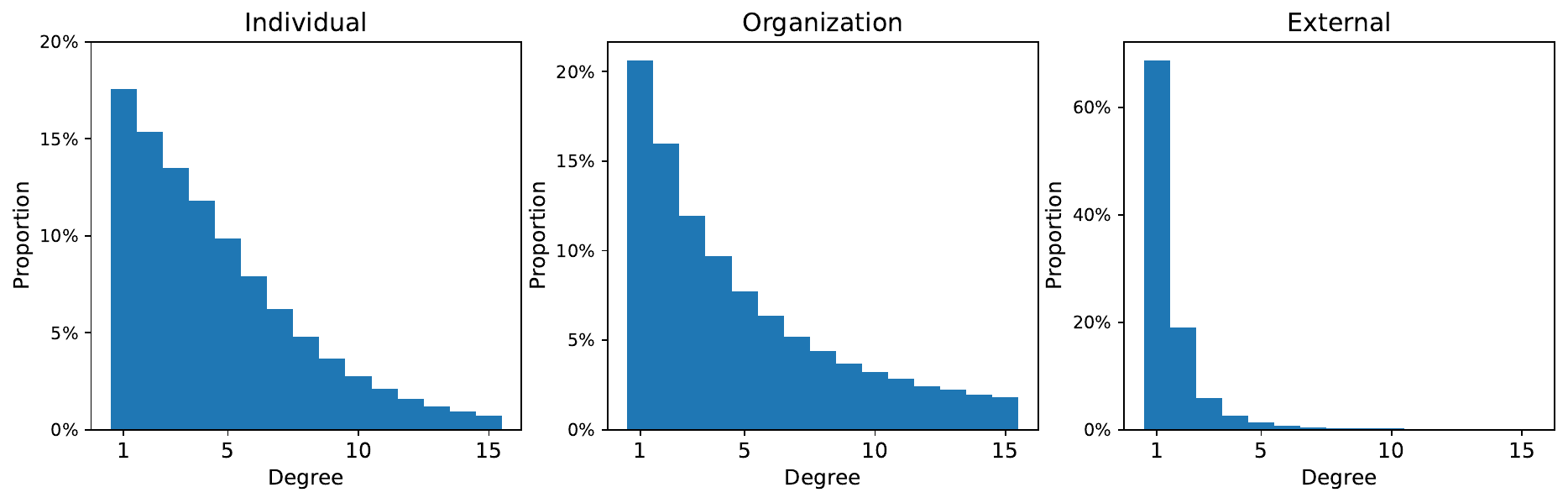}
\caption{Histogram showing the distribution of the degree centrality for the three different node types, completely ignoring the edge type.}
\label{fig:degree_hist}
\end{figure}

Due to the sensitive nature of these data, containing both personal and possibly competition sensitive information for the bank, the data are not shareable. 
We are neither allowed to reveal the exact details of the graphs nor the features associated with the different nodes/edges in our model. 
Below, we give a broad overview of the characteristics and features of the graph, within our permission restrictions. 
To get a feeling of the local characteristics of the graph, Figure \ref{fig:egonet} shows egonets of four random nodes, with the starting node enlarged. 
The upper and lower panels show, respectively, the 3-hop and 9-hop egonets of the (undirected) transaction and role edges. The number of shown hops was chosen to balance presentability and amount of detail.
Moreover, Figure \ref{fig:degree_hist} shows histograms of the degree centrality for the three different node types.
Some nodes have a large number of neighbors, while others have few. 
While the degree distribution of individuals and organizations is similar, the distribution for organizations has a thicker tail, indicating that it's more common for organizations to exhibit a higher degree.
As we don't have knowledge of edges between external nodes, this node type typically has much fewer neighbors.


The three node types have separate sets of features that contain basic information about the entity.
There are eleven, eight, and two node features for, respectively, individuals, organizations, and external nodes.
The transaction edges have as features the number and monetary amount of transactions made within the one-year period. 
The role edges have two features, the first denoting the role type and the second the ownership percentage (provided the role represents ownership in the organization).

\subsection{Experiment setup}

The goal of this use case and experiment is to see to what extent our HMPNN method is able to predict the label (suspicious/regular) on nodes where the label is unknown to the model.
As mentioned in the Introduction, we concentrate on building models for detecting fraudulent \textit{individual} customers.
This means that even if we use the entire graph for message passing, only individual nodes (in the training set) are assigned a label to learn from, and only individual nodes (in the test set) are subsequently predicted and evaluated. 

To evaluate the performance of our methods, we benchmark and compare their performance to a set of alternative models/methods. 
To mimic a scenario with unknown labels, we split the nodes into a training set and a test set. 
The whole graph is available for each of the models when training them, but the labels are only available for the nodes in the training set. 
At testing time, each method attempts to predict the (unknown) nodes in the test set, and their performance is compared using different performance measures.
We used a 70-30 train-test split, where the splitting was performed using stratified random sampling with allocation proportional to the original class balance, such that the class balance is preserved in the two sets. 
The same split was used for all methods.

We compare the result of four model frameworks: two non-graph methods supplied with additional node features, and two GNN methods. 
The models are applied with multiple model complexity configurations. 
To enhance the non-graph methods, we generated 83 additional node features that accompany the original node features. These include 11 summaries of the node-neighborhoods, i.e.~the number of neighbors of different types, both incoming and outgoing. We also computed 8 weighted summaries of node-neighborhoods, specifically for transaction edges and the monetary amount node feature. 
Additionally, we added 64 features generated using metapath2vec. These were assembled from embeddings of dimension 8 generated for each meta-path of length 2 starting and ending at a node of type individual. 
Together with the 11 intrinsic node features, this results in a total of 94 features. For more detailed information on these additional node features, please refer to Appendix \ref{app:network_features}.
The four methods are briefly described below:
\begin{description}
\item[Logistic regression] This classic method serves as a very basic benchmark not directly utilizing the network information and with a basic and inflexible parametric form. The method has access to the additional network-generated node features.
\item[Regular Neural Network] This method, applied with both 1 and 2 hidden layers, is much more flexible than the logistic regression model, but neither utilizes the network directly. The method has access to the additional network-generated node features.
\item[HGraphSage] GraphSage \citep{HAMILTON2017}  is a well-known homogeneous GNN method and is applied to our heterogeneous graph in the same fashion as HMPNN, as described below. In contrast to MPNN, GraphSage does not utilize edge features. We, therefore, test its performance both with and without additional node features that hold the weighted in/out degrees for the transaction edges, and the weight is the transaction amount on the edges. This results in 6 additional node features for nodes of type \textit{individual} and \textit{organization}, and 4 on nodes of type \textit{external}. We test the method with the aggregation function in \eqref{eq_mpnn_agg_sum} (HMPNN-sum). We applied the method with both one, two, and three hidden layers. 
\item[HMPNN] This GNN method, described in Section \ref{sec:HMPNN}, is our extension of MPNN to heterogeneous graphs. We test the method with the two aggregation functions in \eqref{eq_mpnn_agg_sum} (HMPNN-sum) and \eqref{eq_mpnn_agg_ct} (HMPNN-ct).
We applied the method with both one, two, and three hidden layers for each of the aggregation methods.
\end{description}
In total, our experiment contains 15 different models/method variants. 
The number of parameters involved in each of these is listed in Table \ref{tab:num-parameters} in the Appendix.

%



The logistic regression and Regular Neural Network models were trained using the open source deep learning library Pytorch \citep{NEURIPS2019_9015}. 
HMPNN and HGraphSage were trained using the open source library Pytorch Geometric (PyG), which expands Pytorch with utilities for representing and training GNNs. 

All models were trained with the Adam optimiser \citep{Kingma2015Adam}, using the Binary Cross Entropy loss function:
\[
\text{Loss}(\hat{y}_v, y_v) = y_v\cdot \log(\hat{y}_v) + (1-y_v)\cdot \log(1-\hat{y}_v).
\]
The hyperparameters for each of the methods were tuned using 5-fold cross-validation on the training set. 
Here, three hyperparameters were determined: (1) The regularization strength,  (2) the learning rate, and (3) the number of training iterations, i.e., by early stopping. 
For all methods, the learning rate was in the range $[10^{-4},10^{-1}]$. 
As for regularization, the $L_2$-constraint was used, and was in the range $[10^{-8},10^{-1}]$. 
The optimal value for the $L_2$ constraint was highly dependent on the complexity of the model to be trained.

The experiments were carried out using  \textit{python 3.7.0}, with PyTorch 1.12.1 and PyG 2.2.0.
The computer used to run the experiments had 8 CPUs of the type \textit{High-frequency Intel Xeon E5-2686 v4 (Broadwell) processors}, with 61GB shared memory, and one GPU of type \textit{NVIDIA Tesla V100} with 16GB memory. 
This GPU has 5,120 CUDA Cores and 640 Tensor Cores.

\subsection{Results}

\begin{figure}[ht]
    \centering
    \includegraphics[width=1\textwidth]{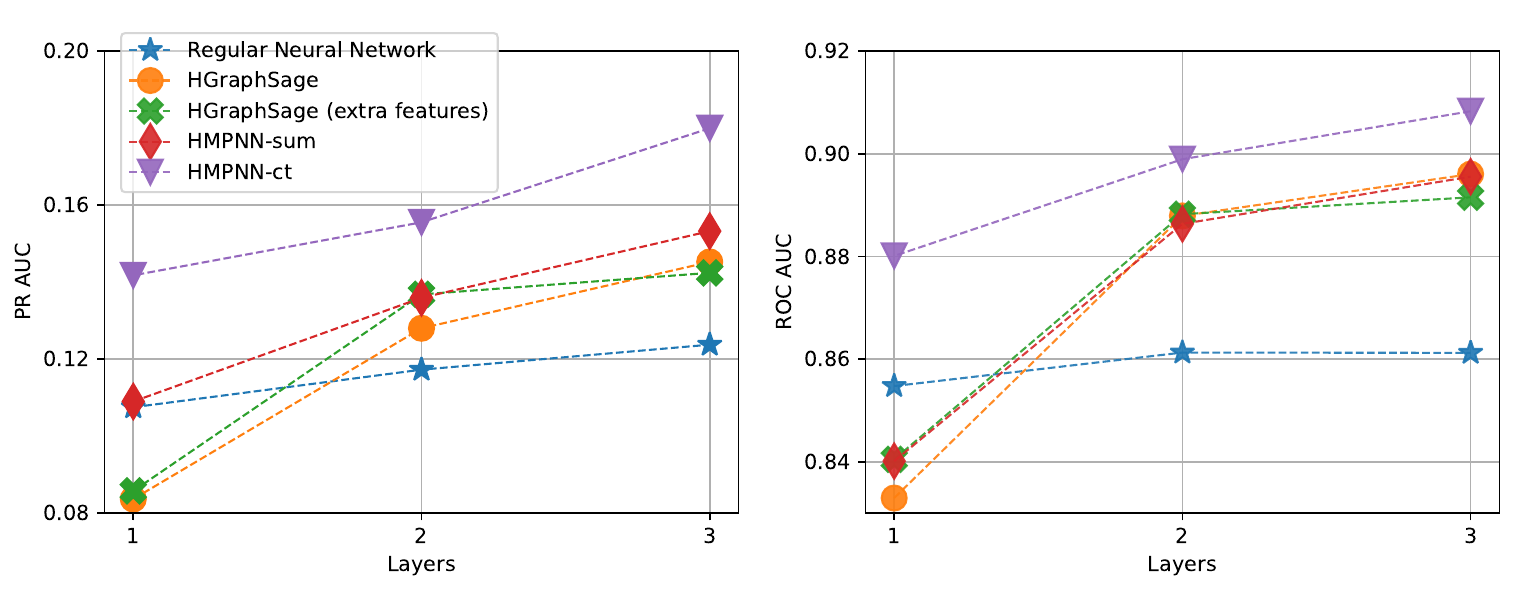}
    \caption{ROC AUC AND PR AUC on the test set for all different methods and number of neural network layers used by the methods. The Regular Neural Network with one layer corresponds to Logistic Regression.}
    \label{fig:AUCres}
\end{figure}

For each node in the test set, all the different methods output a score between 0 and 1, reflecting the probability that the node is a suspicious customer. 
To measure the overall performance of different methods on the test set we rely on the area under the precision/recall curve (PR AUC) and the area under the receiver operator curve (ROC AUC). 
PR AUC computes the area under the curve obtained by plotting the \textit{precision} (TP/(TP+FP)) as a function of the \textit{recall} (TP/(TP+FN)), and PR ROC computes the area under the curve obtained by plotting the recall as a function of \textit{False Positive Rate} (FP/(FP+TN)). Here TP/FP/TN/FN represents the number of classified nodes which are, respectively, true positives, false positives, true negatives, and false negatives. 
Figure \ref{fig:AUCres} displays the ROC AUC and PR AUC on the test set for all different methods and number of neural network layers used by the respective methods. 

All methods benefit from including more layers. 
HMPNN-ct with 3 layers is the best method in terms of both PR AUC and ROC AUC.
While HMPNN-ct does quite well when applied with a single hidden layer, this is not the case for the other network models, at least when compared to the basic logistic regression (regular neural network with one layer). 
In terms of ROC AUC, the logistic regression model is actually better than all the other network models, and in terms of PR AUC, it is better than the HGraphSage models and comparable to HMPNN-sum. 
This is quite remarkable as the logistic regression model does not know anything about the network, except for additional network summary features, and has the simplest form of architecture. 
Note, however, that the Regular Neural Networks with 2 and 3 layers only do slightly better than Logistic Regression. This indicates that the simple architecture of the Logistic Regression model is not a significant downside for that limited data set.
 Moreover, the large performance gap to the HMPNN-ct model also shows that it is certainly possible to get more out of the network structure than the other network models manage. Thus, we believe that the lack of performance for the other network models is related to an inappropriate and inefficient architecture compared to that of HMPNN-ct. 
 The fact that overall, HMPNN-sum does not perform on par with HMPNN-ct further indicates that the performance boost in HMPNN-ct is mainly due to the architectural trick of the last single-layer neural network. 


\begin{table}[t]
\centering
\small
\caption{Precision at specific values of Recall, in addition to PR AUC and ROC AUC for the different models}
\label{tab:precision-recall}
\begin{tabular}{p{2.5cm}ccccccc}
\toprule
\multirowcell{2}{Model}&Number & \multicolumn{4}{c}{Recall(\%)} & PR & ROC \\ \cline{3-6}
&of Layers & 1 & 5 & 10 & 50 & AUC & AUC \\ 
\hline
\multirowcell{3}{Regular\\Neural Network}
&1 & 61.54 & 36.28 & 28.55 & 5.53 & 0.1075 & 0.8547 \\
&2 & 64.00 & 43.82 & 30.51 & 5.89 & 0.1173& 0.8613\\
&3 & 61.54 & 47.56 & 34.68 &5.88 & 0.1237 & 0.8612\\
\hline
\multirowcell{3}{HGraphSage}
&1& 59.26 & 33.62 & 20.58 & 4.01 & 0.0836 & 0.8329 \\
&2& 61.54 & 48.15 & 38.56 & 6.51 & 0.1280 & 0.8879 \\
&3& 59.26 & 60.47 & 42.47 &7.19 & 0.1452 & 0.8960 \\
\hline
\multirowcell{3}{HGraphSage\\(extra features)}
&1& 48.48 & 34.21 & 21.29 & 4.24 & 0.0858 & 0.8405 \\
&2& 64.00 & 50.65 & 35.39 & 7.41 & 0.1368 & 0.8882 \\
&3& 69.57 & 50.32 & 38.27 &7.68 & 0.1424 & 0.8915 \\
\hline
\multirowcell{3}{HMPNN-sum}
&1& 64.00 & 38.24 & 29.75 & 5.34 & 0.1090 & 0.8401 \\
&2& 69.57 & 42.62 & 36.38 & 7.76 & 0.1359 & 0.8863 \\
&3& 84.21 & 53.79 & 38.27 &8.67 & 0.1532 & 0.8955 \\
\hline
\multirowcell{3}{HMPNN-ct}
&1& 76.19 & 48.75 & 38.18 & 6.92 & 0.1418 & 0.8801 \\
&2& 61.54 & 50.65 & 43.66 & 8.96 & 0.1555 & 0.8989 \\
&3& 66.67 & 58.21 & 50.99 &10.25 & 0.1800 & 0.9083 \\
\bottomrule
\end{tabular}
\end{table}

Considering the limitations in resources faced by banks, conducting thorough examinations of a substantial volume of suspicious cases is typically unfeasible.
Therefore, the primary purpose of the model is to generate a limited set of  high-quality predictions where money laundering is likely to occur, meaning that the precision at small to medium-sized recall levels is more relevant than the higher ones.
Table \ref{tab:precision-recall} shows the precision corresponding to recall levels of 1\%, 5\%, 10\% and 50\%, respectively, and allows studying the performance of the methods in greater depth and at a wider range.

Focusing on HMPNN-ct, we see that when the classification threshold is set such that we identify 1\% of the suspicious customers (recall = 1\%) two-thirds of those classified as suspicious \textit{are} actually suspicious (precision $\approx 67\%$). 
Increasing the classification threshold to 5\% or 10\% gives precisions of about 58\% and 51\%.
Moreover, if we decrease the threshold such that half of the suspicious customers (recall = 50\%) are identified, 90\% of the customers classified as suspicious are not really suspicious. 
These rates may not seem impressive at first glance. 
Considering the severe class imbalance in the data (less than 0.5\% of the total number of observations are suspicious), and the fact that detection of money laundering is a notoriously difficult problem, these performance scores are, actually, very promising. 

Finally, note that even though the 3-layer HMPNN-sum model performs worse than HMPNN-ct overall and for the larger recalls, it obtains a significantly better precision (84\% vs 67\%) at recall 1\%. In essence, this model is better at detecting the most evident instances of money laundering. 
This aspect is crucial to consider when selecting a model, particularly if resource limitations restrict the investigation to a small number of customers for potential money laundering.

\section{Summary and concluding remarks}
\label{sec:conclusion}

The present paper proposed and applied a heterogeneous extension of the homogeneous GNN model, MPNN \citep{GILMER2017}, to detect money laundering in a large-scale real-world heterogeneous graph. 
The graph is derived from data originating from Norway's largest bank, encompassing 5 million nodes and close to 10 million edges, and comprises customer data, transaction data, and business role data. 
Our heterogeneous MPNN model (HMPNN) incorporates distinct message-passing operators for each combination of node and edge types to account for the graph's heterogeneity. 
Two versions of the model are proposed: HMPNN-sum and HMPNN-ct. 
Notably, HMPNN-ct used a novel strategy for constructing the final node embeddings, as the embeddings from each node-edge operator were concatenated and fed into a final single-layer neural network. Overall, this version outperformed HMPNN-sum as well as all alternative models by a significant margin. HMPNN-sum was, however, the best model at recall = 1\%, i.e.~it was most accurate for the customers assigned the largest probabilities of being suspicious.

As we saw from the overall measures in Figure \ref{fig:AUCres}, all models performed the best when fitted using 3 hidden layers.  We may have gotten even better performance if we increased the number of layers further. However, 3 layers is where we hit the memory roof on our GPU, so we were unable to explore this in practice.
It is worth noting that the HMPNN-ct architecture is also clearly the most successful when we are restricting the number of layers to 2 or 1. This is relevant as larger networks, and/or less computational resources or training time available, may in other situations demand a reduction in the number of layers. From Table \ref{tab:num-parameters} in the Appendix, we also see that apart from the regular neural network model, HMPNN-ct has the fewest number of model parameters of the 3-layer models, indicating that it has an efficient architecture.

Customers labeled as ``regular'' may, in fact, be suspicious and potentially involved in money laundering -- they just haven't been controlled in the existing AML system and are therefore labeled as ``regular''. This scenario holds true for practically all instances of money laundering modeling.
Consequently, if a customer labeled as "regular" is assigned a high probability of being suspicious by the model, it is possible that the customer has been mislabeled.
 As a result, modeling test phases with a labeled test set, as outlined here, can also serve as a means to generate suggestions for customers who warrant further investigation into their past behaviors.
In other words, the modeling approach allows for the identification of customers who may require re-examination based on the model's predictions, even if they were initially labeled as "regular".

When implementing a predictive model for suspicious transactions within a real AML system, several crucial decisions need to be made. 
One vital consideration involves determining the optimal stage in the process (see Figure \ref{fig:alert_workflow}) for applying the predictions: either preceding the alert inspection or preceding the case investigation.
If the predictions are used prior to the alert inspection, it might be preferable to set a classification threshold with a higher recall. On the other hand, if the predictions are applied before the case investigation, it may be more appropriate to select a more stringent threshold, i.e.~that has a lower recall. This would help minimize the allocation of investigation resources towards false positives, leading to greater efficiency.

In order to increase the performance of our money laundering modeling approach, some aspects become readily apparent. 
While our data set is rich in terms of the number of nodes and edges, contains network data from both financial transactions and professional roles, and has a large number of edge and node features, it can always be richer.
In our setup, the transaction edges contain the number and total monetary amount made in the one-year period. This could be refined by also including the standard deviation, median, and other summary statistics like in \citet{JULLUM2020}. 
Moreover, provided high-quality data can be obtained, it would be valuable to include customer links using connections from social media platforms, geographical information such as shared address or phone numbers, or even family relations. 

As mentioned, organizational customers are fewer in number and exhibit less homogeneity compared to individuals, rendering them less suitable for modeling compared to individuals.
It still presents a natural candidate for further work to develop models that make predictions on the bank's organization customers. 
However, to truly see the potential of such a model, we believe that it is essential to expand the dataset in time to include more fraudulent organizations to learn from and enrich the data with more organization-specific features.

A significant limitation that applies to our work, as well as most endeavors related to money laundering detection, is the restricted nature of the data, which is confined to customers within a single bank. Although our graph includes "external" customers from other banks, the transactions and professional role links between customers in external banks are unavailable. This is primarily due to banks being either unwilling or prohibited from merging their customer information with other banks, due to security regulations or competitive considerations.
Surmounting these data-sharing challenges among prominent financial institutions would enable the modeling of a more comprehensive network of transactions and relationships, leaving fewer hiding places for money launderers. Nevertheless, the administration of such collaborative, analytical, and modeling systems demands substantial resources and investments, a process that may take years to accumulate.



In any case, to the best of our knowledge, no scientific work has been published regarding the utilization of heterogeneous graph neural networks in the context of detecting money laundering on a large-scale real-world graph. 
This paper should be viewed as a first attempt to leverage heterogeneous GNN architecture within AML and has showcased promising outcomes.
We envision that our paper will provide invaluable perspectives and directions for scholars and professionals engaged in money laundering modeling. Ultimately, we hope this contribution will aid in the continuous endeavors to combat money laundering.

\section*{Code availability}
The implementation of our HMPNN model is available here: \url{https://github.com/fredjo89/heterogeneous-mpnn}

\section*{Declaration of competing interest}

The authors declare that they have no known competing financial interests or personal relationships that could have appeared to influence the work reported in this paper.

\section*{Acknowledgements}

Funding: This work was supported by the Norwegian Research Council [grant number 237718].
\appendix

\section{Network features} 
\label{app:network_features}
This appendix provides details of how we created additional node features that capture information about a node's role in the network. 
These were utilized in our entity-based models (and HGraphSage) for benchmarking in Section \ref{sec:usecase}.

We generated a total of 83 additional features, which when combined with the original 11 intrinsic node features, resulted in a total of 94 features. These additional features are categorized into three distinct types: 1) Unweighted Neighborhood summary consisting of 11 features, 2) Weighted Neighborhood summary consisting of 8 features, and 3) Metapath2vec embeddings with a total of 64 features.

\textit{Unweighted Neighborhood Summary}.
The unweighted neighborhood summary features encapsulate the (unweighted) in/out degree for each meta-step that nodes of type individual are part of, see Figure \ref{fig:graph_schema}. This amounts to seven features, six from transaction edges and one from role edges.
Further, four summary features are added:
1) The total in-degree, which is the count of all incoming edges, irrespective of the meta-step,
2) The total out-degree, which is the count of all outgoing edges, irrespective of the meta-step,
3) The total degree, which is the count of all edges, irrespective of the meta-step or the direction,
4) The total count of meta-steps in which the node is involved.
In total, this gives 11 additional node features.

\textit{Weighted Neighborhood Summary}.
The weighted neighborhood summary features contain the weighted in/out degree for each meta-step involving transaction edge. Here the edge feature representing the monetary amount was used as edge weight. This provides six features for a node of type individual.  
Further, two summary features are added:
1) The total weighted in-degree, 
2) The total weighted out-degree, 
In total, this gives 8 additional node features.

\textit{Metapath2vec Embeddings}.
Metapath2Vec was used to generate embeddings for each of the four meta-paths shown in Table \ref{tab:metapath2vec_metapaths}. 
These are all meta-paths of length 2 that start and end at a node of type individual. 
The dimension of the embedding for each meta-path was set to 8. 
The embeddings were added as features on both the node at the start and end of the meta-path. This amounts to 64 additional node features. 
We used the implementation of MetaPath2Vec in Pytorch Geometric. Table \ref{tab:MetaPath2Vec_settings} lists the parameters used in the creation of the embeddings.  

\begin{table}[ht!]
    \centering
    \small
    \caption{Meta-paths for Individuals}
    \label{tab:metapath2vec_metapaths}
    \begin{tabular}{c}
\hline
$\texttt{ind} \overset{\texttt{txn}}{\xrightarrow{\hspace*{1.5cm}}} \texttt{ind} \overset{\texttt{txn}}{\xrightarrow{\hspace*{1.5cm}}} \texttt{ind}$ \\
$\texttt{ind} \overset{\texttt{txn}}{\xrightarrow{\hspace*{1.5cm}}} \texttt{org} \overset{\texttt{txn}}{\xrightarrow{\hspace*{1.5cm}}} \texttt{ind}$ \\
$\texttt{ind} \overset{\texttt{txn}}{\xrightarrow{\hspace*{1.5cm}}} \texttt{ext}\overset{\texttt{txn}}{\xrightarrow{\hspace*{1.5cm}}} \texttt{ind}$ \\
$\texttt{ind} \overset{\texttt{role}}{\xrightarrow{\hspace*{1.5cm}}} \texttt{org} \overset{\texttt{txn}}{\xrightarrow{\hspace*{1.5cm}}} \texttt{ind} $ \\
\hline
    \end{tabular}
\end{table}

\begin{table}[ht!]
    \centering
    \footnotesize
    \caption{Hyperparameters for the creation of MetaPath2Vec-embeddings.}
    \label{tab:MetaPath2Vec_settings}
    \begin{tabular}{c|c}
    \hline
    Parameter & Value \\
    \hline
    \texttt{embedding\_dim}                 &8\\
    \texttt{walk\_length}                   &20\\
    \texttt{context\_size}                  &10\\
    \texttt{walks\_per\_node}               &10\\
    \texttt{num\_negative\_samples}         &1\\
    \hline
    \end{tabular}
\end{table}

\section{Network parameters} 
\label{app:parameters}

Table \ref{tab:num-parameters} shows the number of parameters for each of the models discussed in Section \ref{sec:usecase}. 

\begin{table}[ht!]
\centering
\small
\sisetup{group-separator={,},group-four-digits=true}
\caption{Number of model parameters for the different models}
\label{tab:num-parameters}
\begin{tabular}{lSSS}
\toprule
\multicolumn{1}{c}{\multirow{2}{*}{Model}} &  \multicolumn{3}{c}{Number of Layers}  \\ 
\cmidrule(r{5pt}){2-4}
& \multicolumn{1}{c}{1} & \multicolumn{1}{c}{2} & \multicolumn{1}{c}{3}  \\
\midrule 
NeuralNetwork & 95 & 9025 & 17955 \\
HGraphSage & 189 & 3999 & 7809 \\
HGraphSage (extra features) & 329 & 13045 & 25761 \\
HMPNN-sum & 296 & 6536 & 12776 \\
HMPNN-ct & 3071 & 4487 & 6303 \\
\bottomrule
\end{tabular}
\end{table}

\clearpage
\bibliographystyle{model5-names}
\bibliography{bibliography}
\end{document}